\documentclass[10pt,letterpaper]{article}

\usepackage{cogsci}
\usepackage{pslatex}
\usepackage{apacite}
\usepackage{amsmath}
\usepackage{amssymb}
\usepackage{graphicx}
\usepackage{pgfplots}
\usepackage{tikz}
\usepackage[ruled,vlined]{algorithm2e}

\usepackage{color}

\cogscifinalcopy % Uncomment this line for the final submission 

\title{Improving Multi-Agent Cooperation using Theory of Mind}

\author{{\large \bf Terence X. Lim$^{1}$ (terence.lim@u.nus.edu{})}, {\large \bf Sidney Tio$^{2}$ (sidney.tio@aisingapore.org)}, \\ {\large \bf Desmond C. Ong$^{1,3}$ (dco@comp.nus.edu.sg{})} \\
  {}$^{1}$Department of Information Systems and Analytics, National University of Singapore \\
  {}$^{2}$AI Singapore \\
  {}$^{3}$Institute of High Performance Computing, Agency for Science, Technology and Research (A*STAR), Singapore}

\begin{document}

\maketitle

% \begin{verbatim}https://www.overleaf.com/project/5da9cdf0d1b3da000142849c
% Outline: ToM in cooperative game playing
% model
% experiments ToM-ToM; ToM-noToM; noToM-noToM;
%             ToM-human; noToM-human; (human-human)
% results
% ideally: noToM-noToM < ToM-noToM < (noToM-human) < 
%          ToM-ToM < (ToM-human) < human-human]
% do we add in deep RL?
% 1st page: Intro
% p.2-3: Model
% p.3-4: Experiment
% p.4-5: Results
% p.6: Discussion, references
% \end{verbatim}

\begin{abstract}
Recent advances in Artificial Intelligence have produced agents that can beat human world champions at games like Go, Starcraft, and Dota2. However, most of these models do not seem to play in a human-like manner: People infer others' intentions from their behaviour, and use these inferences in scheming and strategizing. Here, using a Bayesian Theory of Mind (ToM) approach, we investigated how much an explicit representation of others' intentions improves performance in a cooperative game. We compared the performance of humans playing with optimal-planning agents with and without ToM, in a cooperative game where players have to flexibly cooperate to achieve joint goals. We find that teams with ToM agents significantly outperform non-ToM agents when collaborating with all types of partners: non-ToM, ToM, as well as human players, and that the benefit of ToM increases the more ToM agents there are. These findings have implications for designing better cooperative agents.

%Partially Observable Markov Decision Processes (POMDP) seek to model an agent's decision process, taking into account the agent's uncertainty about the environment. Inspired by recent Theory of Mind (ToM) approaches, we desire to model the agent's ability to infer the intentions of others in a multi-agent environment using inference in a nested POMDP.
%Given observations in a particular state, we aim to use interactive planning with POMDPs on the agent to infer goals and subsequently predict the actions of the other agents in a cooperative setting. We will be performing experiments that model agents in a gaming environment, specifically using games which involves 2 or more players who share a common goal but with unassigned subgoals. We hypothesize that ToM reasoning will benefit agents in a setting with more agents and more task complexity. We will use simulations of I-POMDP agents to demonstrate the utility of this approach.

\textbf{Keywords:} 
Bayesian Theory of Mind; Multi-Agent Games; Partially Observable Markov Decision Process
\end{abstract}

\section{Introduction}

%\hspace{0.4cm}

In recent years, we have seen tremendous advancements in Artificial Intelligence (AI) that have produced agents that can beat human world champions at competitive games such as Go \cite{silver2016mastering}, StarCraft II \cite{vinyals2019grandmaster}, and Dota2 \cite{openai2019dota}. Indeed, researchers have now produced agents that can solve many aspects of complex games once thought to be well outside the realm of artificial intelligence, including games involving imperfect information (Hanabi; \citeNP{lerer2020search}), hidden player roles (Avalon: The Resistance; \citeNP{serrino2019finding}), and even negotiations \cite{lewis2017deal, mell2018results}.

% "keeping track of what seems to work and what doesn't" ?

Despite this impressive progress, the vast majority of these AI agents do not seem to be playing these games in the same manner that humans do. Consider Deep Reinforcement Learning (RL), which was used to train many of these impressive game-playing feats. RL agents start off \textit{tabula rasa} with no knowledge of rules or strategies, and learn to play the game via trial-and-error in self-play; instances of agents are pitted against each other, and the actions used by the more successful agents are integrated into tried-and-tested strategies. The introduction of deep neural networks (and better search methods) to learn complex non-linear approximations of the ``value" of states and actions have enabled deep RL agents to tackle increasingly complex games in recent years \cite<e.g., >{silver2018general, carroll2019utility}. However, while human players also learn by trial-and-error, they learn much faster than deep RL agents. People also, by contrast, start off with more complex representations of rules and strategies---e.g., whether another social agent is ``helping" or ``hindering" \cite{ullman2009help}---which consequently help humans to learn more quickly and generalize better than the current state of modern AI, given the same amount of game-play experience. Although there is no denying the performance of deep RL agents at these complex games, one has to wonder if, by incorporating principles of human learning and reasoning into these models, we can further improve these agents and perhaps overcome some of their present limitations.

One promising approach to a more human-like game-playing agent lies in explicitly modelling intentions. Recent work in the Bayesian Theory of Mind \cite{baker2017rational} framework has modelled how laypeople infer the intentions and preferences \cite{baker2009action}, intuitive rewards and costs \cite{jara2016naive}, knowledge states \cite{baker2017rational}, social goals \cite{ullman2009help}, and even emotions \cite{ong2019computational} and utterance pragmatics \cite{goodman2016pragmatic} of other agents from their behaviour. In this computational cognitive modelling framework, based on folk belief-desire psychology \cite<e.g.,>{wellman1990simple}, \textbf{agents} are assumed to be rational, utility-maximizing agents who choose their actions to maximize their utility given their latent preferences and knowledge about the world. Then, by observing an agent's actions, an independent \textbf{observer} can infer the agent's underlying motivation, such as intentions, by using Bayes' rule. The work thus far in Bayesian ToM have mostly been focused on explaining human observer inferences---i.e., modeling how people infer the intentions of other agents---and they have not yet studied how these inferences can then be incorporated into a decision-making model and used to improve coordination. We note also that computer scientists have studied such nested reasoning models using Partially Observerable Markov Decision Processes \cite<POMDPs, e.g., >{doshi2009monte, gopalan2015modeling}, but they have not empirically compared them with human performance. Here, we address these gaps by (i) implementing a Bayesian ToM-inspired POMDP model in a collaborative setting with humans, and (ii) empirically comparing the performance of humans with agents with and without a ToM.

In this work, we investigate how well an explicit representation of intentions benefit cooperation. We chose a game where players work together without communication to maximize team score, but where the task allocation is fluid. That is, Tasks $A$, $B$, and $C$ are required to accomplish a goal, but any of the agents can accomplish any of the tasks. Importantly, we hypothesize that in such an environment with fluid and flexible task allocation and in the absence of explicit communication, agents with an explicit ToM would be significantly better partners than optimal planners with no ToM.

To illustrate, consider the task of two (or more!) chefs preparing a meal without any form of communication, head chef giving orders, and without any prior history working together\footnote{Of course, one solution to coordination without communication is extensive training with Standard Operating Procedures, as is done in many high-stakes domains like the military, fire-fighting, and high-end restaurant kitchens. Here, we do not address pre-training.}. Vegetables need to be collected and chopped; chopped vegetables need to be cooked; cooked meals need to be served. A little Theory of Mind---along with several assumptions that everyone is cooperative and optimal---can go a long way: Bob is standing closest to the vegetables, and everyone can reason that Bob can reach the vegetables the earliest, and so Bob may be the optimal choice to prepare them. Bob then starts walking to the vegetables, which other agents can then take as a confirmation that indeed, Bob is preparing the vegetables---in fact, Bob \textit{knows} that the others know that Bob is taking this task. The other agents can then proceed to address other tasks. In such a manner, seamless cooperation may emerge naturally out of rational, cooperative agents with a Theory of Mind, even in the absence of communication.

\begin{figure*}[htp]
    \centering
    \includegraphics[width=1.0\textwidth]{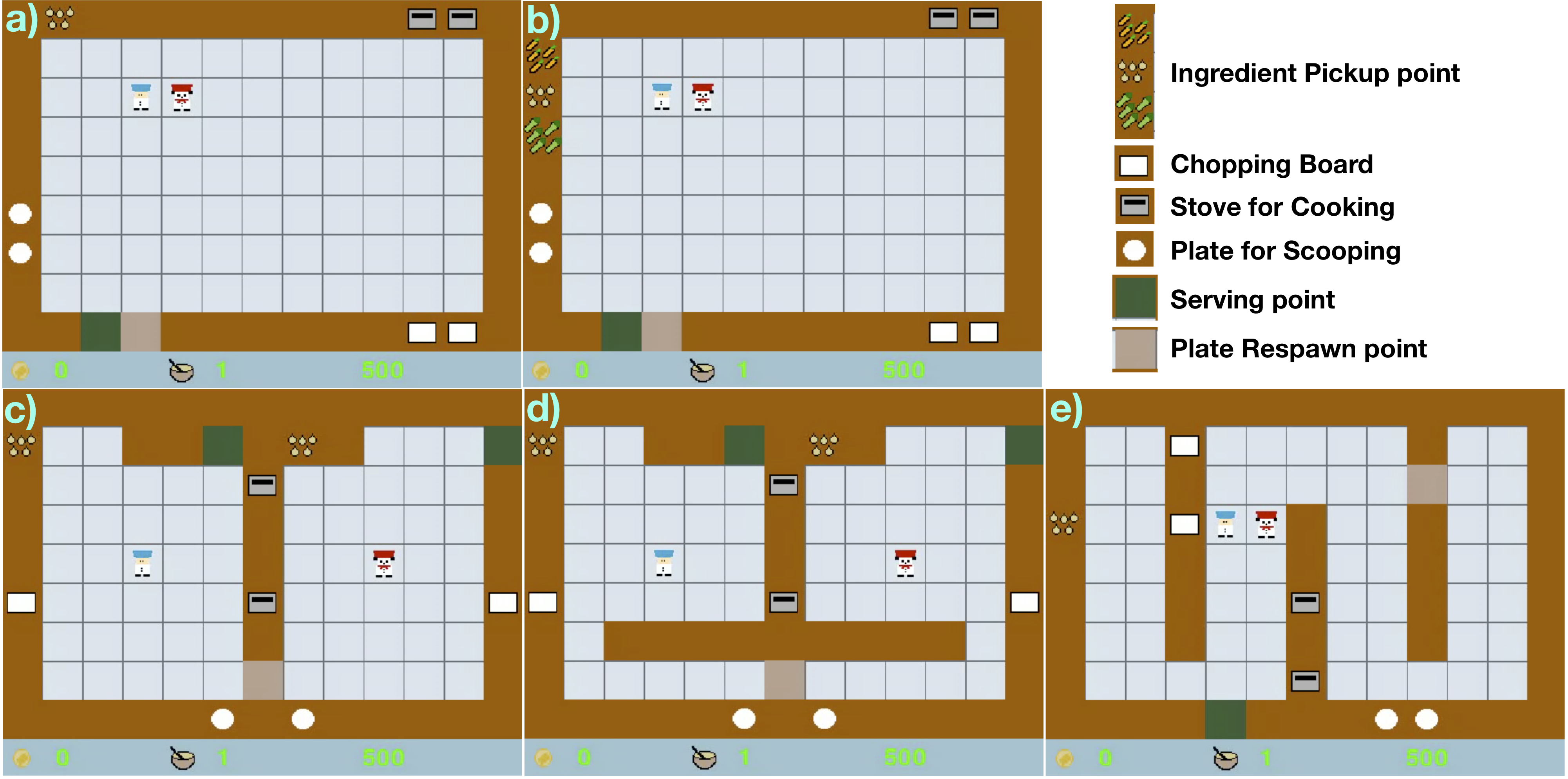}
    %\caption{Experimental layouts. \textit{Open-Space}; map (a) and (b). \textit{Asymmetric Advantages}; map (c) and (d). \textit{Counter Circuit}; map (e).}
    \caption{Experimental layouts. Maps (a) and (b) provide large open spaces with tasks located at the boundaries. Maps (c) and (d) give asymmetric task advantages to each player. Map (e) introduces a winding circuit that increases the traveling cost to various tasks, making it even more important to minimize redundancy.}
    \label{fig:exp_layout}
\end{figure*}

To test this idea, we chose to build an environment inspired by the popular multiplayer party game, Overcooked\footnote{Overcooked is a 2016 cooking simulation game developed by Ghost Town Games and published by Team17.}. Players (or our AI) each control a chef in a kitchen, and have to prepare specific meals, earning points for each successful dish served (Fig. \ref{fig:exp_layout}). We note that this same game was also recently and independently studied with deep RL agents \cite{carroll2019utility}, which we return to in the discussion. We tested teams of human players along with AI with and without a Theory of Mind, and we find that ToM agents are more effective teammates than non-ToM agents. Furthermore, ToM agents seem to be more effective on maps where there are large distances between subtasks, and hence larger effort-savings by inferring others' intentions. Finally, simulations with more than two agents also showed that ToM agents work even better in teams with other ToM agents.

\section{Bayesian Theory of Mind in cooperation}

In this section, we lay out the design of a rational planner agent with no Theory of Mind (nToM), as well as its counterpart with a ToM. For both, we use a Markov Decision Process (MDP) formulation. The nToM agent completely observes the world state but is oblivious to other agents' intentions, while the ToM agent models other agents' intentions as a partially-observable component of the world. A Partially-Observable MDP consists of $\langle\textit{S}, \textit{A}, \textit{T}, \textit{R}, \Omega, \textit{O}\rangle$, where
\begin{itemize}
    \setlength\itemsep{-0.5em}
    \setlength\itemindent{2em}
    \item \textit{S} denotes a finite set of states of the physical environment (i.e., $x,y$ positions of each agent, goal locations)
    \item \textit{A} denotes a finite set of actions, i.e., movement in eight directions (up, up-right, right, ...) and five `task actions' (pick up, drop, chop, cook, serve).
    \item %$\textit{T}: \textit{S} \times \textit{A} \mapsto {\Pi(\textit{S})}$
    $\textit{T}$ is the state-transition function, where $\textit{T}(s, a, s')$ gives the probability of ending in state $s'$, given that the agent starts in state $s$ and executes action $a$.
    \item $\textit{R}: \textit{S} \times \textit{A} \mapsto \mathbb{R}$ is the reward function, providing the immediate reward gained by the agent for taking each action in each state.
    \item \textit{$\Omega$}: is the finite set of latent intentions the observer infers from the world and observed actions.
    \item $O: S \times A \mapsto \Pi(\Omega)$ is the observation function, i.e., the inference function, which specifies a probability distribution over possible inferences, given the current world state and the current actions.
\end{itemize}

\begin{figure}[htp]
    \centering
    \includegraphics[width=0.9\columnwidth]{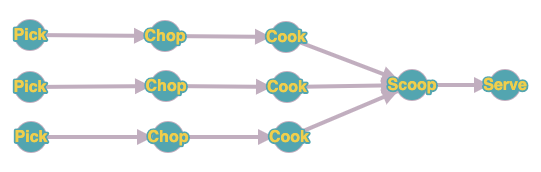}
    \caption{Directed Graph showing the subtask dependencies for the goal of serving a Soup}
    \label{fig:main_goal_dag}
\end{figure}

\textbf{Decomposing goals into subtasks.} In four of our five maps, we used the same simple recipe: one needs to pick, chop, and cook three (3) onions in the same pot, and then serve the Onion Soup to get points. (In the fifth map, we had a mixed vegetable soup where instead of 3 onions, there are 3 different vegetables). Thus, the goal of serving a Soup can be broken down into several sub-tasks, which have their own pre-requisite conditions. For example, an agent has to be holding a raw, unchopped ingredient and standing next to the chopping board before they can ``Chop". We structured these sub-tasks as a Directed Acyclic Graph (Fig. \ref{fig:main_goal_dag}) to specify the dependencies between the sub-tasks to accomplish the goal.

\textbf{Rational Planner Agent (nToM).} The rational planner agent (without a ToM) plans optimally: It knows all the shared goals, such as how many soup orders are pending, as well as the state of all possible subtasks. Agents earn points only for completing the entire goal, so we added internal rewards to incentivize agents to prioritize completing each order before going to the next one, i.e., finishing subtasks leading to one goal before starting the next goal. We chose to give more internal rewards to `later' subtasks: 10 for \textit{picking} an ingredient, 20 for \textit{chopping}, 40 for \textit{cooking}, 50 for \textit{scooping} and 100 for \textit{serving}. These internal rewards are not added to the total score, and are used solely for subtask prioritization.

Agents compute the utility of all subtasks $g$ as the internal reward $R'(g)$, less the cost of completing that subtask from the current state $s$: $c(g | s)$. We used a cost of 1 per step, and no cost for actions (\textit{pick}...). Agents then choose subtasks proportional to their utility, via a softmax decision-rule:
\begin{align}
P(g|s) \propto \exp\left[ \beta (R'(g) - c(g|s)) \right] \label{eqn:softmax_choice}
\end{align}
where the optimality parameter $\beta$ controls agent behaviour: agents approach strictly rational behaviour as $\beta \to \infty$, and completely random behaviour when $\beta = 0$. Here, we set $\beta=0.5$. Finally, after choosing a subtask, agents then use A* Search as a heuristic best-first search to find the shortest path(s) to that subtask, giving $P(a|g,s)$.

% To account for possible deviation from rational agents' ideal behaviour, we perform goal weighting for the current episode given the observed action taken in the previous episode, by sampling goals with probability proportional to their exponential expected-utility. More specifically, their expected utility is calculated using the softmax function with a $\beta$ parameter [0, $\infty$], where the agent achieves stricter rational behaviour as $\beta$ \to $\infty$. Thus, while agents tend to prioritize goals that maximize their expected-utility, tweaking the $\beta$ parameter allows them to sometimes choose a non-optimal goal. 

% considers all possible goals
% internal rewards - steps to get there
% internal reward shaping: prioritize completing each order before going to the next one, more internal rewards to 'later' goals in the DAG.
% we used a cost of -1 per step.
% internal rewards: pick 10 chop 20 cook 30 scoop 40

\textbf{Theory of Mind Agent.} The ToM agent model proceeds in two steps. First, it infers what the other agent's possible intentions are. Second, it then incorporates the other agent's intentions into its own planning model. We note that we only consider one level of nesting here \cite{baker2017rational}; that is, ToM agents reason about other agents only, and do not recursively reason about other agents reasoning about themselves.

Given the world state at the previous time-step $s$, and other agents' last action $a$, the ToM model infers the posterior of agents' subtasks $g$ using Bayes' rule:
\begin{equation}
    P(g|a,s) = \frac{P(a|g,s)P(g|s)}{P(a|s)} = \frac{P(a|g,s)P(g|s)}{\sum_{g'\in{G}}P(a|g',s)P(g',s)}
    \label{bayeseqn}
\end{equation}

We used as a non-informative prior\footnote{We considered using expected utility as in Eqn. \ref{eqn:softmax_choice} to model the prior $P(g|s)$, but chose to leave it uninformative as a conservative choice for modelling human players.}, a uniform distribution over $P(g|s)$. To solve this posterior inference, the ToM model would, for each other agent, sample each subtask a hundred times, and for each subtask compute the most-likely actions that the agent would take $P(a|g,s)$. The model then conditions on which action was actually observed to compute $P(g|a,s)$. Finally, to allow for a non-zero probability for unlikely actions, we used Laplace smoothing and added one count to each possible action.

The next step after inferring other agents' subtasks, is coordination planning: that is, using the other agents' subtasks to coordinate one's own actions. The ToM agent ``plans" from the perspective of other agents and calculates their expected utilities of each subtask (Reward $-$ Cost to get there). In the event where the ToM agent's chosen subtask $g'$ returns higher reward for another agent, it would then choose to abandon subtask $g'$. Interestingly, given that all agents receive the same subtask rewards, this boils down to comparing relative distances to the subtasks, and so this decision-rule can also be explained by a simple heuristic: If both the ToM agent and another agent have the same subtask $g'$ in mind \textit{and} the other agent is closer to $g'$, then the ToM agent will abandon $g'$, assuming that the other agent will take care of it. Basically, if my partner is closer, they have it covered, and I can move onto another task. If the ToM agent abandons $g'$, it will then remove $g'$ from its list of available subtasks, and then re-plan for the next subtask. By avoiding these conflicts, ToM agents would then be able to reduce redundant steps (e.g., both chasing after the same subtask when only one is needed) during collaborative goal pursuit.

\section{Experiment}

We designed five maps for our cooperative game to vary in difficulty and with certain characteristics that we predicted would impact cooperation (Fig. \ref{fig:exp_layout}). The first two maps, Fig. \ref{fig:exp_layout} (a) and (b), %present low-level coordination challenges, as they 
provide vast, open spaces where subtasks are relatively far from each other. Maps (c) and (d) provide asymmetric subtask advantages for each player. For example, for the left agent, the distance from the cooking pots to the serving point is shorter. Finally, Map (e) has a winding circuit that increases the distance to certain tasks, such as picking ingredients, which incentivizes taking fewer redundant steps. The map also allows more sophisticated cooperation strategies: For instance, one agent could pick and chop the onion (1 step), while the other agent could pick up the chopped onion from the other side of the chopping board to bring it to the cooking pot (3 steps), compared to the first agent doing everything (8 steps)\footnote{Post-hoc qualitative analysis suggested that some humans pairs do learn and successfully apply this particular strategy, but none of our AI agents (could have) learnt this.}.

%From the maps shown in Fig. \ref{fig:exp_layout}\cite{carroll2019utility}, \textit{Open-Space};map (a) & (b) presents low-level coordination challenges, where, in this vast open space, tasks are relatively far away from each other and it is easy for agents to avoid wasting steps to get to a certain task. \textit{Asymmetric Advantages};map (c) and (d) challenges agents to coordinate and play to their strengths since certain sides require less effort in completing certain goals; \textit{Counter Circuit};map (e) challenges agents to pass ingredients over counters to save steps and specialize in performing a certain goal instead of carrying items around.

We recruited N=111 (Mean age = 22.58, 49.5\% female) participants, who were randomly assigned to three conditions: \textbf{Human-nToM} (N=27 individuals), \textbf{Human-ToM} (N=28 individuals), and \textbf{Human-Human} (N=28 pairs). After giving consent, participants were shown a screenshot of the game and were briefed about the objectives and the key commands, which were also indicated on the keyboard. Participants then played a short training round of 150 time-steps to familiarize themselves with the game: each time-step corresponds to taking 1 step in the game. Following this, participants played all 5 maps, each for 500 time-steps. The order in which participants played the maps was randomized across participants. Participants in the \textit{Human-Human} condition were not allowed to discuss any strategies or even communicate: verbal exclamations, eye-contact, and gestures were prohibited. This was to ensure the fairness of comparing results across the other conditions since our AI agents were neither able to communicate with humans nor with themselves.

To compare the performance of the Human conditions with all-agent conditions, we additionally ran simulations for three more AI-only conditions: \textbf{nToM-nToM} (N=30 runs), \textbf{ToM-nToM} (N=30 runs), and \textbf{ToM-ToM} (N=30 runs). The number of runs were chosen to match the sample size in the Human conditions, rounding off to 30.

\section{Results}

\begin{figure}[htp!]
    \includegraphics[width=1.0\columnwidth]{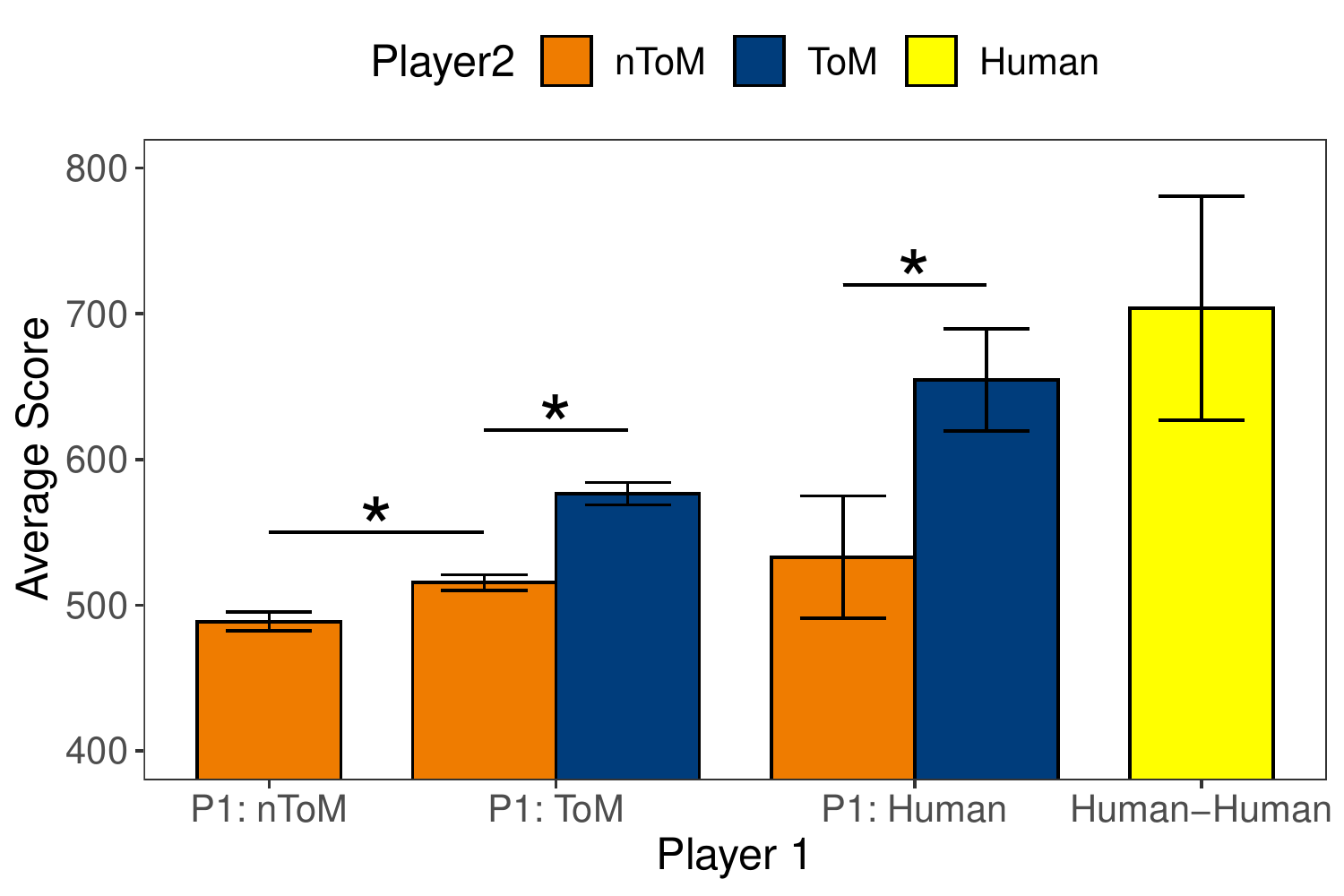}
    \caption{Average score by condition. Error bars represent between-subject 95\% Confidence Intervals averaged over maps, which visually under-estimates the between-condition effects from a repeated-measures analysis. * indicate significant \textit{a priori} predicted comparisons, all $p<.001$.}
    \label{fig:main-results}
\end{figure}

We present the averaged scores by conditions in Fig. \ref{fig:main-results}. As we expected, participants in the \textit{Human-Human} condition scored the highest, while agents in the \textit{nTom-nToM} condition performed the worst amongst all the conditions. For reference, successfully serving one dish the instant it appears earns players 150 points, but this decreases by 1 per time-step (so serving a dish 50 time-steps after the order appears will earn 100 points). This is also the only way to earn points.

We expected that ToM agents would be better cooperators and hence produce higher results than nToM agents. This is formalized in three \textit{a priori} hypotheses, all contrasting a ToM agent with a nToM agent: [1] \textit{Human-ToM} $>$ \textit{Human-nToM}, [2] \textit{ToM-ToM} $>$ \textit{ToM-nToM}, and [3] \textit{ToM-nToM} $>$ \textit{nToM-nToM}. To test these hypotheses, we made dummy variables that corresponded to these specific contrasts (e.g., a dummy that is 1 if \textit{Human-ToM} and 0 if \textit{Human-nToM} for contrast [1], and we looked only at the subset of data in these two conditions), and fit linear mixed-effects models with random intercepts by player and by map, to control for the repeated-measures nature of the data. As we expected, all of the comparisons were significant. ToM agents were better cooperators than nToM agents when playing with humans (comparison [1], $b=121.8$ $[$95\% CI: $69.7, 173.9],$ $t=4.58, p<.001$), playing with other ToM agents (comparison [2], $b=60.9$ $[50.2, 71.6],$ $t=11.2, p<.001$), or playing with other nToM agents (comparison [3], $b=26.8$ $[16.5, 37.1],$ $t=5.1, p<.001$). For visual reference, these comparisons are indicated with an $*$ in Fig. \ref{fig:main-results}, going from right to left.

%We present quantitative results for different agent pairings in Fig.\ref{fig:main-results}. With the innate ability of humans to learn over time using Theory of Mind, it was no surprise that \textbf{Human-Human} experiments achieved the best results. The lowest performance threshold was set by non-ToM and non-ToM simulation runs where the agents performed the necessary tasks with no regards of each other's behaviour. Even so, there is a large gap between the experiments between humans and AI agents. Utilizing the comparison of a ToM-agent and human (Fig. \ref{fig:main-results}; middle columns) playing with both non-ToM agent and ToM agent showed significant difference in performance when thrown into a collaborative setting. This supports our hypothesis that ToM-agents reason well in cooperative settings and would perform more optimally and avoid making moves that may cause confusion of their intentions to human players. The difference in performance by Non-ToM agents vs. ToM-agents are further elevated as the complexity of the environment (maps with optimization strategies like passing over the counter) and tasks involved (multiple ingredient recipe task) increases.

\begin{figure*}[htp!]
    \includegraphics[width=0.9\textwidth]{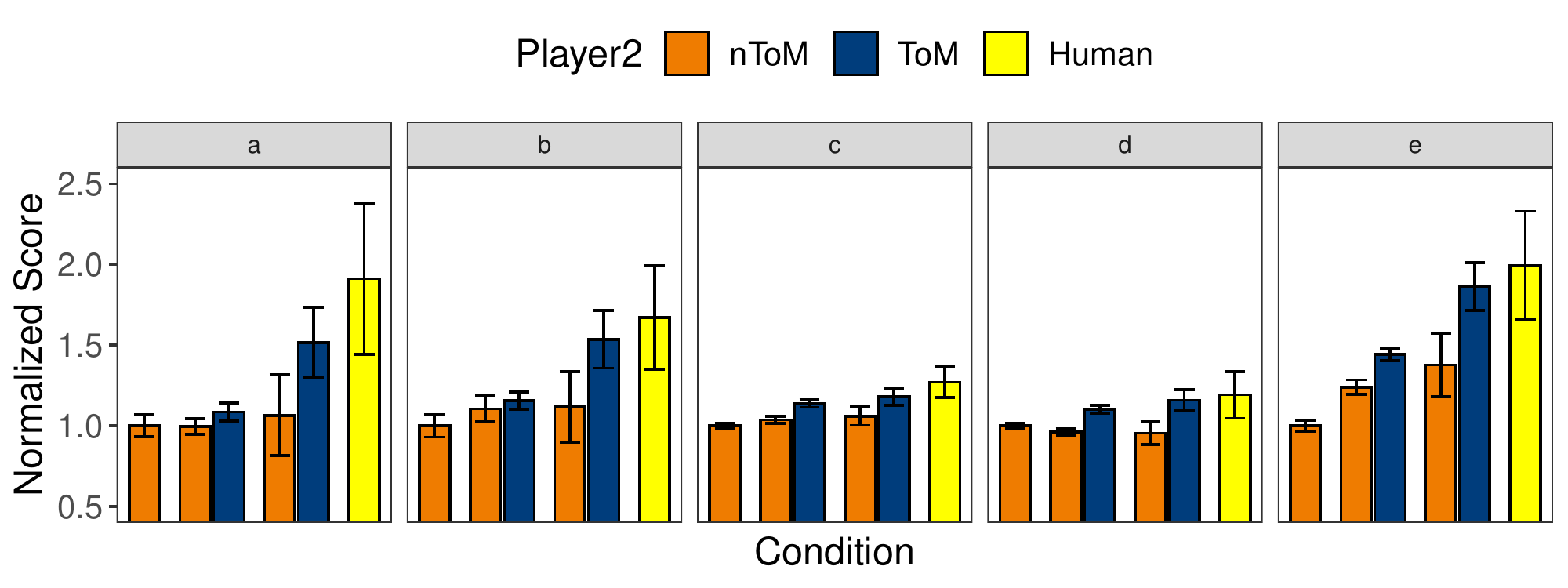}
    \caption{Normalized score by condition, for each map separately. The arrangement of the bars are identical to Fig. \ref{fig:main-results}; Please refer to Fig. \ref{fig:main-results} for the horizontal axis labels, which are omitted for clarity. Scores are normalized such that \textit{nToM-nToM} is 1.0, to better illustrate the replication of the pattern of results in Fig. \ref{fig:main-results}. Error bars represent between-subject 95\% CI.}
    \label{fig:results-allmaps}
\end{figure*}

\textbf{Map features.} Next, we examined performance on each map. There were large differences in mean performances across the maps: mean \textit{nTom-nTom} performance was lowest for maps (a) [131 points] and (b) [200 points], then map (e) [420 points], (d) [690 points] and (c) [998 points]. 

However, we were most interested in features of the maps that ToM agents were able to exploit, hence we looked at maps with the highest (relative) ToM $>$ nToM difference. In Fig. \ref{fig:results-allmaps}, we plot the normalized score for each map separately, where we normalized all the scores by dividing by the mean \textit{nTom-nTom} performance in that map, so the \textit{nTom-nTom} normalized score is 1.0, and we can examine the ToM benefit as a relative improvement.
We find that the ToM benefit is the largest in maps (a), (b), and (e). We suggest that this could be due to large distances between subtasks, which greatly benefited ToM agents who expend less effort going after subtasks that other agents are also pursuing. Surprisingly, this did not happen in maps (c) and (d) though we expected ToM to be more efficient as well at exploiting asymmetries in the map.

% mean score in $a = b < e < d < c$
% a 131.2333
% b 199.9333
% c 998.4333
% d 692.3000
% e 421.9667

\textbf{Simulations with more than two agents.} Finally, in order to investigate the benefits of a ToM when the number of agents increase, we ran additional simulations with more than two agents, varying the number of ToM agents in the team. Specifically, we ran four different three-agent conditions (\textbf{0-ToM-of-3}, \textbf{1-ToM-of-3}, \textbf{2-ToM-of-3}, and \textbf{3-ToM-of-3}, and five different four-agent conditions (\textbf{0-ToM-of-4}, \textbf{1-ToM-of-4}, \textbf{2-ToM-of-4}, \textbf{3-ToM-of-4}, and \textbf{4-ToM-of-4}). We ran 9 simulation runs per map for each three-agent condition, and 2 runs per map for each four-agent condition. 
%We used the same maps, and we appropriately increased the rate of task orders coming in, so that there are enough subtasks to keep the agents busy.

\begin{figure}[tb!]
    \includegraphics[width=1.0\columnwidth]{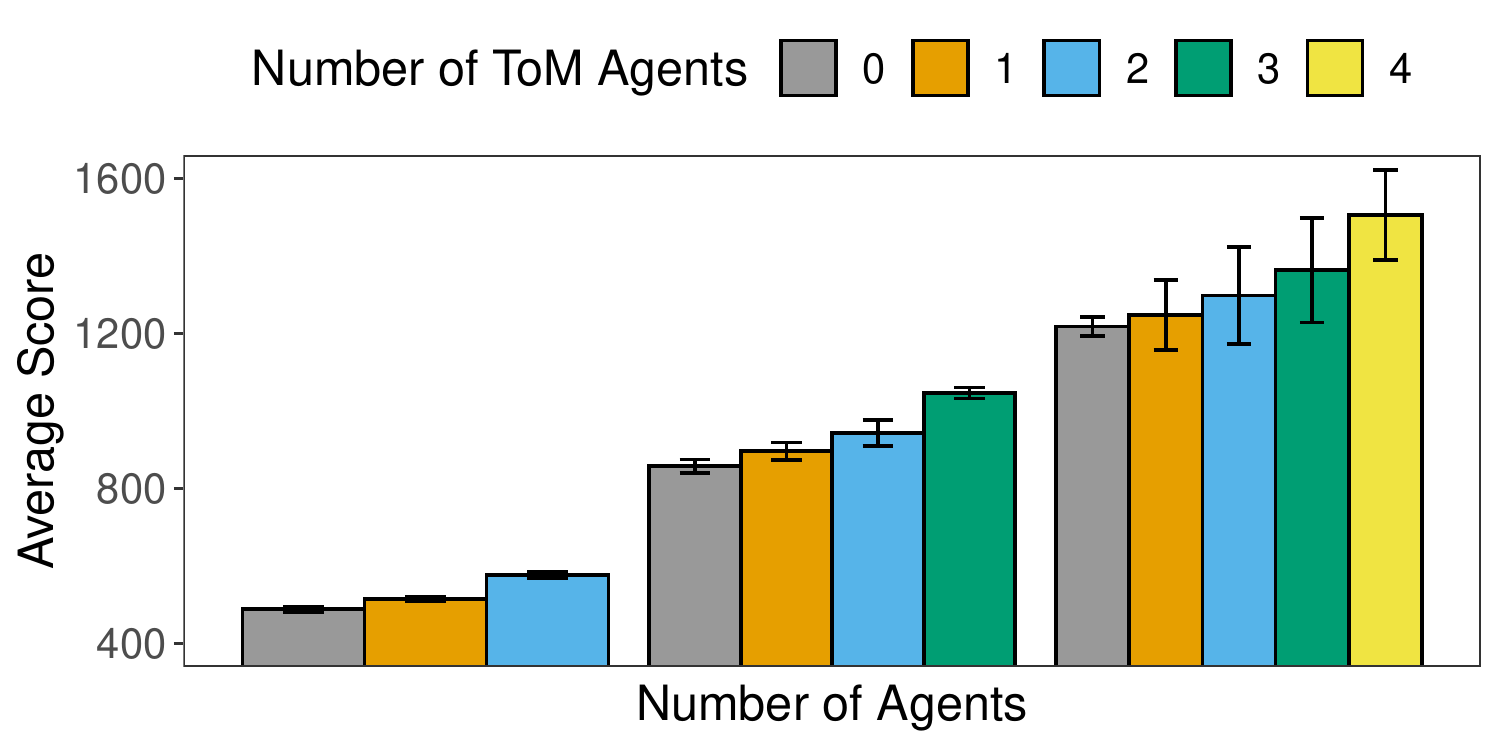}
    \caption{Simulation results for 2-agents (left), 3-agents (middle) and 4-agents (right) games. Colors on adjacent bars represent increasing number of agents with ToM.}
    \label{fig:results-more-agents}
\end{figure}

We show the results from these multi-agent simulations in Fig. \ref{fig:results-more-agents}. From the graph, and from running a linear-mixed effects model with random effects by map, we notice two main effects, which are both expected and highly significant. Increasing the number of agents increases the team score $(b=378$ $[364, 392],$ $t=51.9, p<.001)$, and, for the same number of agents in the team, increasing the number of ToM agents also increases the team score $(b=53.8$ $[45.8, 61.9],$ $t=13.1, p<.001)$. 

In running these simulations, we were interested in whether ToM agents actually benefit teams more when there are already ToM agents, as we hypothesized that ToM agents are better able to cooperate with other ToM agents. This led us to test for an interaction between the number of agents and the number of ToM agents. Indeed, we found a significant positive interaction $(b=15.7$ $[3.7, 27.7],$ $t=2.49, p=0.011)$. This suggests that imbuing each rational planner agent in the team with a Theory of Mind has a \textit{super-additive}, synergistic effect: the more agents there are that reason about other agents' goals, the more effectively they can work together.

% two main effects model [numbers not the most up to date]
%num_agents   375.087      7.387  588.000  50.779   <2e-16 ***
%num_tom       50.330      4.401  588.000  11.435   <2e-16 ***

% interaction model [numbers not the most up to date]
% score ~ num_agents * num_tom + (1 | map_id)
%num_agents          351.523     11.999  587.000  29.295   <2e-16 ***
%num_tom              12.976     15.655  587.000   0.829   0.4075    
%num_agents:num_tom   15.360      6.180  587.000   2.485   0.0132 *  

\section{Discussion}

In this paper we empirically investigated the benefit of adding an explicit Theory of Mind to rational planning agents in a cooperative environment where the allocation of subtasks, needed to accomplish a goal, to agents is fluid. In such an environment, where any agent can accomplish any subtask, a ToM allows more efficient cooperation by reducing redundancy: if I see a teammate walk towards subtask X, I infer their intentions, assume they can handle it, and now I can turn my attention to other subtasks. We showed that ToM agents improve team scores when they play with humans, ToM agents, and non-ToM agents, and that this benefit increases when the environment has large distances between subtasks, translating to larger effort-savings with a ToM. Finally, we also showed via simulations of more than 2 agents, that the benefits of ToM scale synergistically; ToM agents contribute even more effectively when there are already other ToM agents on the team.

It is impressive that we find these strong results even though our study has much room for improvement. First, our model had several parameters (e.g., softmax $\beta$, internal rewards for reward shaping) that were not optimized to data. Second, we used only one level of ToM reasoning, although previous studies in humans have shown that one to two levels is sufficient, and it may not be computationally-efficient to recurse until convergence. Finally, ToM agents do not act knowing that others are reasoning about them---we used a simple rule for coordination planning that had ToM agents give up subtasks that gave higher utility to and were likely to be accomplished by others. One can imagine more sophisticated reasoning that involves ToM agents acting \textit{on the assumption} that other agents similarly reason about their intentions. Despite this not being made explicit in our models, we still find a synergistic benefit of ToM agents working with other ToM agents.

We motivated this paper by reviewing how much progress deep RL agents have made in game-playing. Indeed, \citeA{carroll2019utility} recently and independently studied a similar environment, and proposed several RL training regimes to train agents to play with humans. We think that deep RL and ToM rational planner approaches have much to offer each other, and they are in fact, complementary rather than competing. Our current agents cannot learn new strategies like RL (such as dropping ingredients on counters to pass them to other agents), but can play (well) with humans without any training. Future work should look at integrating deep RL and ToM approaches to leverage the advantages of each approach.

In sum, there is much utility from building social, cooperative agents that can better reason about human partners. Our work shows that an explicit representation of intentions can dramatically boost performance in a dynamic context where subtask allocation is fluid.

%\cite{albrecht2018autonomous}

%Inferring intention from language / communication

%other types of situations where ToM may be beneficial.

\section{Acknowledgments}

%Acknowledgements redacted for blind review.
% Acknowledge AI SG Grant? Startup
This research is supported by the National Research Foundation Singapore under its AI Singapore Programme (Award Number: AISG-RP-2019-011), and by a Singapore Ministry Of Education Academic Research Fund Tier 1 grant.

\bibliographystyle{apacite}

\setlength{\bibleftmargin}{.125in}
\setlength{\bibindent}{-\bibleftmargin}

\bibliography{ToM}

\end{document}